\documentclass[preprint,12pt]{elsarticle}

\usepackage{amssymb}
\usepackage{amsfonts,amssymb}
\usepackage[utf8]{inputenc}
\usepackage{graphicx}
\usepackage{caption}
\usepackage{multirow}
\usepackage{booktabs}
\usepackage{amsmath}
\usepackage{amsthm}
\usepackage{subfigure}
\usepackage{algorithm}
\usepackage{algorithmic}
\usepackage{mathrsfs}
\usepackage{diagbox}

\graphicspath{ {./figure/} }

\journal{Journal of Franklin Institute}

\begin{document}

\begin{frontmatter}



%\title{Safe Exploration based Reinforcement Learning for Navigating Intersections in Dense Traffic}

\title{Multi-task Safe Reinforcement Learning for Navigating Intersections in Dense Traffic}


\author[inst1,inst2]{Yuqi Liu}

\affiliation[inst1]{organization={The State Key Laboratory of Management and Control for Complex Systems, Institute of Automation, Chinese Academy of Sciences},%Department and Organization
%            addressline={Address One}, 
%            city={City One},
            postcode={100190}, 
            state={Beijing},
            country={China}}

\author[inst3]{Yinfeng Gao}
\author[inst1,inst2]{Qichao Zhang}
\author[inst3]{Dawei Ding}
\author[inst1,inst2]{Dongbin Zhao}
\affiliation[inst2]{organization={College of Artificial Intelligence, University of Chinese Academy of Sciences},
            postcode={100049}, 
            state={Beijing},
            country={China}}
            
\affiliation[inst3]{organization={            School of Automation and Electrical Engineering, University of Science and Technology Beijing},%Department and Organization
 %           addressline={Address Two}, 
 %           city={City Two},
            postcode={1000839}, 
            state={Beijing},
            country={China}}

\begin{abstract}

Multi-task intersection navigation including the unprotected turning left, turning right, and going straight in dense traffic is still a challenging task for autonomous driving. For the human driver, the negotiation skill with other interactive vehicles is the key to guarantee safety and efficiency. However, it is hard to balance the safety and efficiency of the autonomous vehicle for multi-task intersection navigation. In this paper, we formulate a multi-task safe reinforcement learning with social attention to improve the safety and efficiency when interacting with other traffic participants. Specifically, the social attention module is used to focus on the states of negotiation vehicles. In addition, a safety layer is added to the multi-task reinforcement learning framework to guarantee safe negotiation. We compare the experiments in the simulator SUMO with abundant traffic flows and CARLA with high-fidelity vehicle models, which both show that the proposed algorithm can improve safety with consistent traffic efficiency for multi-task intersection navigation.

\end{abstract}



%\item A multi-task safe reinforcement learning framework combined with social attention mechanism is proposed, which enhances the safety and efficiency of autonomous system.
% experiments results better than IV baseline

%\item An innovative design of safety layer for collision avoidance is proposed for the intersection navigation problem.
% multi-task with attention

%\item A set of experiments in SUMO and CARLA simulators are designed, and the result shows that the proposed method has a better performance than competitive methods.


\end{frontmatter}


\section{Introduction}
\label{sec:Introduction}

In recent years, autonomous driving has achieved widespread attention in academic and industry communities. However, there are still plenty of problems in interactive high-conflict traffic scenarios such as ramp merging, narrow street passing, unprotected left turn, and so on. The autonomous agent is required to interact with other traffic participants and choose an appropriate strategy to pass through the intersections safely and efficiently.

For the intersection navigation, there are three distinct planning and control approaches: the rule-based method, end-to-end method, and behavior-aware method. The rule-based method is based on some classical models such as Intelligent Driver Model (IDM) \cite{kesting2010enhanced}, Optimal Velocity Model (OVM) \cite{bando1998analysis}, and so on. In addition, the behavior strategies based on the hand-crafted rules are designed by a case-to-case mechanism, which lacks negotiation skills for high-conflict traffic scenarios and generalization ability to new scenarios. End-to-end control approaches such as imitation learning have also been investigated to obtain the driving policy based on image inputs \cite{huang2019learning, codevilla2018end}. However, Waymo \cite{bansal2018chauffeurnet} claims that pure imitation learning is not sufficient even with 30 million examples, which would get stuck or collide in highly interactive scenarios.

Recently, reinforcement learning (RL) is considered as a feasible method to address these issues. RL methods have been widely used in video games \cite{shao2018learning, shao2018starcraft} and autonomous driving \cite{li2019reinforcement, li2019deep}, the success shows its great potential in resolving complex decision-making problems. By combining with other traffic participants’ intentions implicitly, many behavior-aware RL planning methods are proposed such as context and intention-aware partially observable Markov decision process (POMDP) planning, social-aware deep reinforcement learning planning, and so on. Usually, safety and efficiency are considered in the reward shaping. However, it is still a great challenge for the trade-off between safety and efficiency based on reward engineering. For example, a less aggressive agent will certainly spend more time on a similar task. 

During the RL exploration phase, an agent is required to explore as many as different cases in order to find the near-optimal strategy. However, some of those cases may cause critical harm, especially for some physical systems such as robots or autonomous vehicles. Some safe reinforcement learning (safe RL) methods focus on constraining the exploration of an RL agent in order to avoid unsafe conditions. An optimal correction value of original dangerous action will help with the enhancement in safety without losing much efficiency.

\begin{figure}[hbtp]
    \centering
    \includegraphics[width=15cm]{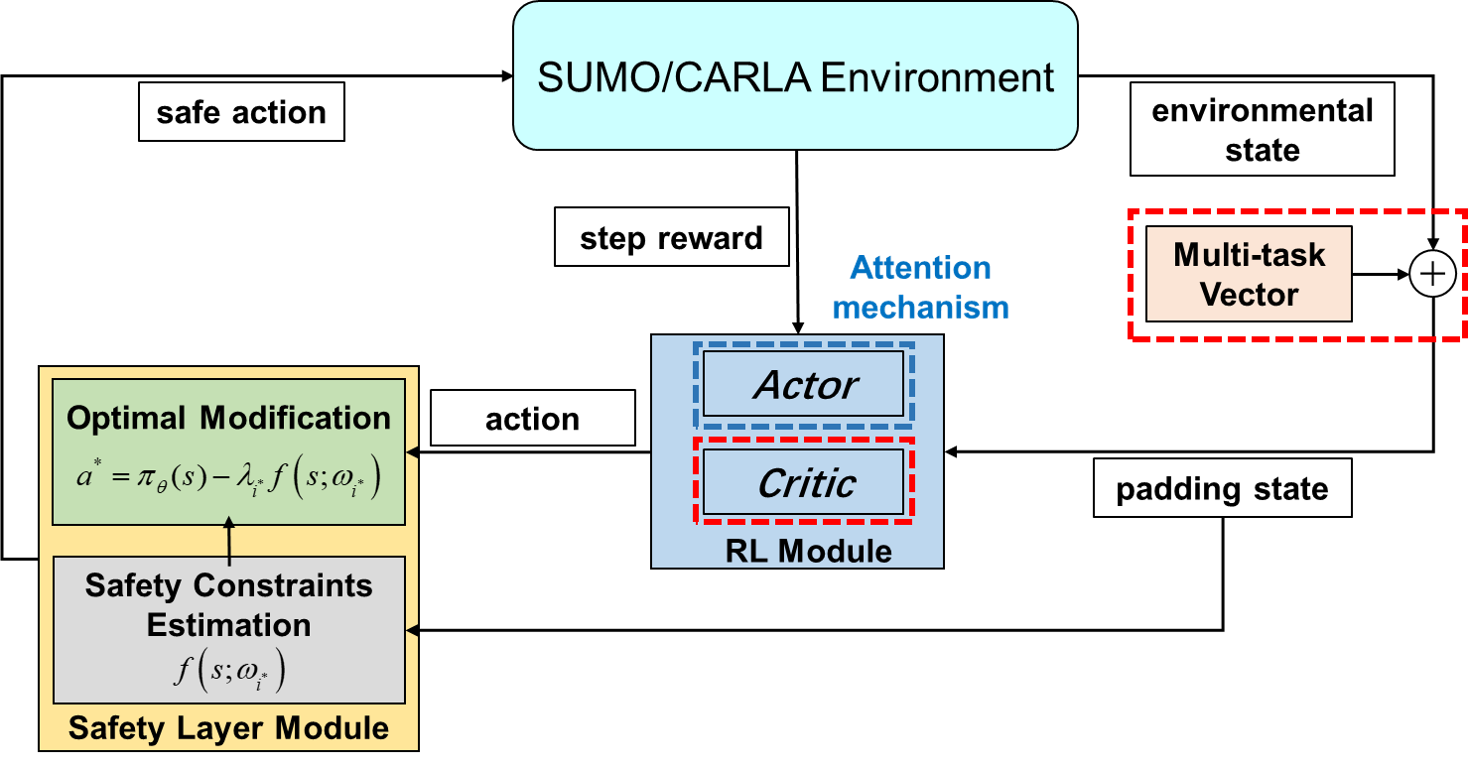}
    \caption{The proposed multi-task safe RL framework}
    \label{Fig: Total framework}
\end{figure}

In this paper, the assumption that ego vehicle in unsignalized intersections
has motion planning information will be hold similarly with \cite{Kai2020Multi,tram2018learning}. We aim to investigate the multi-task unsignalized intersection navigation problem in dense traffic including turning left, going straight, and turning right. To improve the safety and efficiency, a novel multi-task safe RL framework is proposed as shown in Fig.\ref{Fig: Total framework}. Compared with the state of the art (SOTA) work \cite{Kai2020Multi}, the designed safety layer model and actor-critic with social attention can improve negotiation skills of the ego vehicle when interacting with other traffic participants.  As for the original contributions, this paper:
\begin{itemize}
% safety layer design

\item Proposes a multi-task safe RL framework combined with a social attention module, the framework enhances safety and efficiency, also brings better interpretability of state representation;

\item Proposes an innovative design of safety layer for collision avoidance in the intersection navigation problem;
% multi-task with attention

% experiments results better than IV baseline
\item Evaluates the methods with a set of experiments in SUMO and CARLA simulators separately, and the result shows that the proposed method has a better performance than competitive methods.

\end{itemize}

\section{Related Work}
\label{sec:Related Work}

\subsection{Intersection Navigation}
Intersection navigation in dense traffic is one of the most challenging tasks for the autonomous vehicles under urban scenarios, since it is very common to be trapped in the trade-off between safety and efficiency. Recently, a series of RL methods are proposed to solve this problem. \cite{isele2018navigating} proves the effectiveness of deep reinforcement learning in the intersection decision-making problem, and \cite{tram2018learning} improves RL methods’ performance in a similar task by introducing several useful skills to the deep Q-learning(DQN) baseline. \cite{qiao2018pomdp} converts unsignalized intersections navigating task as a hierarchical RL problem, and the hierarchical design of high-level discrete decision and low-level continuous control gains a significant improvement. \cite{bouton2017belief} focuses on the unpredictable characteristics of other traffic participants, introducing the concept of belief state, which improves the ego vehicle’s safety and traffic efficiency. \cite{bouton2019reinforcement} proposes a generic approach to enforce probabilistic guarantees on a RL agent, which constrains acceptable actions of ego vehicle and improves its training efficiency.

% attention + multi-task section
In unsignalized intersection navigation domain, there are other RL methods that resolve the task from different aspects. For example, by integrating social attention mechanism with decision-making progress, a RL agent successfully learns an interaction pattern in \cite{leurent2019social}, which focuses on the social vehicles that are highly related to the ego vehicle’s current state. It gains significant quantitative improvements compared with DQN baseline. In \cite{Kai2020Multi}, unsignalized intersection navigation task is modeled as a multi-task RL problem, in which turning left, turning right, and going straight are considered as specific sub-tasks. Through a multi-task learning framework, the agent learns to handle three navigating tasks at the same time and shows a competitive performance with single-task agents.

\subsection{Safe Exploration}

% introduce the first-order safety model
For the RL tasks in which the safety of agents is particularly concerned, not only that long-term reward maximization is desired, but also damage avoidance is requested. \cite{garcia2015comprehensive} summarizes two major approaches to deploy safe RL methods. The first is to modify the optimality criterion while the second is based on the modification of the exploration process through the incorporation of external knowledge or the guidance of a risk metric. \cite{bouton2019reinforcement} constructs a safe RL method for intersection navigation using linear temporal logic, directly selects safe action within available actions. For more complicated conditions, \cite{bouton2019safe} proposes a method to decompose the scene of intersection navigation, easing the training difficulty of the safe agent.

Though modifications on optimality criterion bring fast convergence in agent training, methods that modify the exploration show a better potential. \cite{isele2018safe} proposes a method by combing a prediction model along with RL training. The prediction model masks unsafe actions to improve the safety performance of an intelligent vehicle. \cite{wen2020safe} proposes a method extending actor-critic frame with an additional risk network to estimate the safety constraint of current policy, while brings a substantial improvement in safety performance. \cite{dalal2018safe} proposes a method to explicitly define a safety constraint in a certain RL environment, and uses a first-order model to estimate the constraint value under an action distribution. According to the constraint model, an analytical solution of optimal safe action can be given. This method is evaluated in a deterministic and non-inertial environment, and it has the potential to be extended into a non-deterministic and inertial problem.

\section{Methods}

\subsection{Definitions}

% define RL and safe RL problem
Here we will use the definition of constrained Markov Decision Processes (CMDP) with a bounded safety signal. We denote by $[K]$ the set $\{1,..,K\}$, and by $[x]^+$ the operation $\max\{x, 0\}$, similarly $[x]^-$ for the operation $\min\{x, 0\}$, where $x\in\mathbb{R}$. A CMDP is a tuple $(\mathcal{S}, \mathcal{A}, \mathcal{P}, \mathcal{R}, \gamma, \mathcal{C})$, where
$\mathcal{S}$ is a state space, $\mathcal{A}$ is an action space, $\mathcal{P}: \mathcal{S} \times \mathcal{A} \times \mathcal{S} \to [0, 1]$ is a transition kernel, $\mathcal{R}: \mathcal{S} \times \mathcal{A} \to R$ is a reward function, $\gamma \in (0, 1)$ is a discount factor, and $\mathcal{C} = \{c_i : \mathcal{S} \times \mathcal{A} \to \mathcal{S} \ |\  i \in [K]\}$ is a set of immediate-constraint functions. Based on that, we also define a set of safety signals $\overline{\mathcal{C}} = \{ \overline{c}_i : \mathcal{S} \to \mathcal{R} | i \in [K]\}$. These are per-state observations of the immediate-constraint values. Policy $\pi : \mathcal{S} \to \mathcal{A}$ refers a stationary mapping from states to actions.

Therefore, a safe RL problem considering an explicit safety constraint can be defined in form of an optimization problem, if all safety signals $\overline{c}_i (\cdot)$ are upper bounded by corresponding constants $C_i \in \mathbb{R}$
\begin{equation}
\begin{array}{c}
\max \limits_{\theta} \mathbb{E}\left[\sum_{t=0}^{\infty} \gamma^{t} R\left(s_{t}, \pi_{\theta}\left(s_{t}\right)\right)\right] \\
\text { s.t. } \quad \bar{c}_{i}\left(s_{t}\right) \leq C_{i},  \forall i \in[K]
\end{array}
\end{equation}

\noindent where $r=R\left(s_{t}, \pi_{\theta}\left(s_{t}\right)\right) $ refers to the reward at timestep $t$, $\bar{c}_{i}\left(s_{t}\right)$ refers to the constraint value of given state $s_t$, $C_{i}$ refers to the upper limit of the constraint value of a given state, $\pi_{ \theta }$ is a parametrized policy.

\subsection{Safety Layer Deployment}

% introduce the safety layer design for sumo experiments
Safety problem in unsignalized intersection scenario is mainly considered as collision risk with other vehicles. All vehicles in the intersection must interact with each other to navigate their own target route. An intelligent agent constructed by deep reinforcement learning will explore all available strategies in order to learn the most proper strategy to solve the problem. In this paper, the safety model takes a two-stage approach to generate a safe action, as shown in Fig.\ref{Fig: Total framework}. Firstly, a safety layer formed by a neural network will implicitly predict safety constraint according to current state. Secondly, the safety model will predict a modification on action conducted by the RL model analytically.

To estimate an impending collision of subsequent timesteps accurately, safety constraint requires a delicate design. Designing of safety constraint is quantization of collision events essentially and can be variable. Besides geometry overlapping, collision can be measured by other means, such as Time to Collision(TTC). During the driving task, a safe RL agent is supposed to satisfy safety constraints.

In this paper, we deploy a neural network trained by offline data for the constraint value estimation. Since the constraint value is defined with domain knowledge, the safety layer is basically a linearization of the dynamics of the state transformation. More specifically, the original safe constraint scalar $\bar{c}_{i}\left(s_{t}\right)$ of a certain timestep can be estimated by the state and action of previous timestep. Therefore, the safety layer model is supposed to estimates the marginal effect of action on the safety constraint value. That is to say that the output of the neural network is the first order derivative of safe constraint value with respect to the action value. Therefore the first-order linearization model can be described as:

\begin{equation}
\bar{c}_{i}\left(s^{\prime}\right) \triangleq c_{i}(s, a) \approx \bar{c}_{i}(s)+f\left(s ; \omega_{i}\right)^{\top} a
\end{equation}

\noindent where $\omega_i$ are weights of the neural network, subscript $i$ refers to the index of constraints if there are multiple ones, $s$ and $s^{\prime}$ refer to the state of two continuous timesteps. We denote $\bar{c}_{i}\left(s^{\prime}\right)$ as $\bar{c_i}'$ for simplicity.

Safety model $f(s; \omega_i)$ takes $s$ as input and outputs a vector which shares the same dimension with $a$. Based on such an assumption, the safety layer model will be trained by solving
\begin{equation}
\mathop{\arg\min}_{\omega_{i}} \sum_{\left(s, a, s^{\prime}\right) \in D}\left(\bar{c}_{i}\left(s^{\prime}\right)-\left(\bar{c}_{i}(s)+f\left(s ; \omega_{i}\right)^{\top} a\right)\right)^{2}
\label{eq3}
\end{equation}
\noindent where $D$ refers to a dataset for model training. With the prediction of slope value $f(s; \omega_i)$ from the safety neural network, the modification of exploration for the original RL can be defined as an optimal problem: 
\begin{equation}
\begin{aligned}
\mathop{\arg\min}_{a} \frac{1}{2}\left\|a-\pi_{\theta}(s)\right\|^{2} \\
\text { s.t. } c_{i}(s, a) \leq C_{i}, \forall i \in[K]  
% \eqno{(1)}
\end{aligned}
\end{equation}
% \noindent Here a typical Euclidean norm loss is used as the optimization target, which refers to that the modified action is supposed to be as close as to the original action. To solve (1) we substitute linear model for $c_i(s, a)$ introduced in (), and get the quadratic program

% \begin{equation}
% \begin{aligned}
% a^{*}= \underset{a}{\arg \min } % \frac{1}{2}\left\|a-\mu_{\theta}(s)\right\|^{2} \\
% \text{ s.t. } \bar{c}_{i}(s)+g\left(s ; \omega_{i}\right)^{\top} a \leq C_{i} \forall i \in[K]
%\end{aligned} 
% \eqno{()}
%\end{equation} 
\noindent An analytical proof has been given by \cite{dalal2018safe}, and the optimal solution is derived by
\begin{equation}
a^{*}=\pi_{\theta}(s)-\lambda_{i^*} f\left(s ; \omega_{i^{*}} \right)
\label{eq5}
\end{equation}

\noindent where 
\begin{equation}
% \begin{aligned}
\lambda_{i}=\left[\frac{f\left(s ; \omega_{i}\right)^{\top} \pi_{\theta}(s)+\bar{c}_{i}(s)-C_{i}}{f\left(s ; \omega_{i}\right)^{\top} f\left(s ; \omega_{i}\right)}\right]^{+},  \\
i^* = \mathop{\arg\max}_{i} {\lambda_i}
% \quad
% i^* = \mathop{\arg\min}_{i} {\lambda_i}^*
% \eqno{}
% \end{aligned}
\label{eq6}
\end{equation}

% \noindent and $i^* = \mathop{\arg\max}_{i} {\lambda_i}$.

The first-order safety above assumes that the constraints values yield a linear relation to the given action, which means the marginal effect of an action value is monotonous at any given time. The structure diagram of the safety layer model is shown in Fig.\ref{Fig: Total framework}.

% % independent safety layer structure
% \begin{figure}[thpb]
%     \centering
%     \includegraphics[width=0.8\linewidth]{safety_layer.png}
%     \caption{Deployment of the safety layer model}
%     \label{safety_model}
% \end{figure}

% safety layer design for intersection navigation
According to the safe exploration method introduced above, a complete procedure can be summarized as follows. At each timestep, the safety layer will predict $f\left(s ; \omega_{i}\right)$ using current state. With the constraint value $\bar{c}_i-C_i$ of current state from the environment, a corrective action will be calculated by (\ref{eq5}) and (\ref{eq6}). Therefore it is critical to define constraints by which the unsafe conditions are reflected accurately. In this paper, we consider collisions between ego and social vehicles as an unsafe condition. Here we propose a safety constraint inspired by the TTC index, which is a critical index in autonomous driving research. 

\begin{figure}[thpb]
    \centering
    \includegraphics[width=0.5\linewidth]{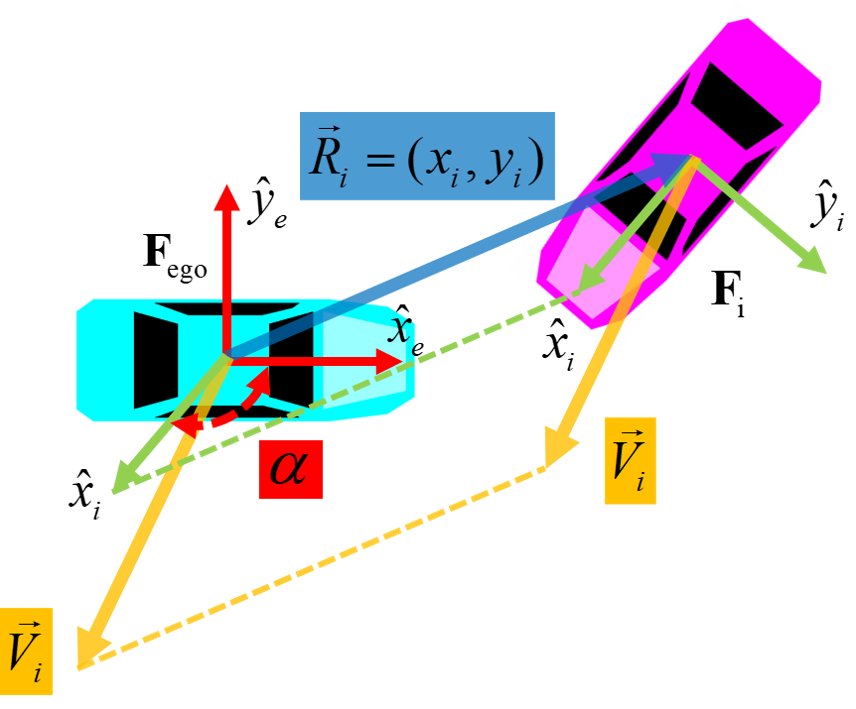}
    \caption{Design of safety constraint inspired by TTC}
    \label{Fig: safety constraint definition}
\end{figure}

The original TTC describes vehicles cruising in a certain lane. Here we use the velocity projected on the relative location vector as an approximation. The instantaneous TTC can be approximately calculated using the relative position vector divided by the relative velocity projection
\begin{equation}
TTC_{fix} = -1 \cdot \dfrac{ | R_i | }{\left [{ | V_i | \cdot \cos{<R_i, V_i>}} \right]^-}
\end{equation}

\noindent where $R_i$ refers to the relative position vector, $V_i$ refers to the relative velocity vector. $|\cdot |$ refers to the 2-norm of a vector. The geometric relationship is shown in Fig.\ref{Fig: safety constraint definition}, $\rm \textbf{F}_{ego}$ and $\rm {\textbf{F}}_{i}$ refer to coordinate frame of ego vehicle and social vehicle respectively, $R_i$ and $V_i$ indicate the relative location vector and relative velocity vector respectively. In order to avoid collision, $TTC_{fix}$ is supposed to be larger than a threshold value, which is as large as several timesteps of simulation. Therefore we deploy the upper limit of constraint $C_i = \frac{t_0}{\eta}$ and $\bar{c}_i=TTC_{fix}$, so that ego vehicle is safe when

\begin{equation} 
\frac{\mu \cdot t_0}{\eta} \leq TTC_{fix}, 0 \leq \eta \leq 1
\label{eq8}
\end{equation}

\noindent where $\eta$ refers to a discount factor, and $t_0$ refers to the timestep length of simulation. So the practical constraint can be written as

\begin{equation}
\bar{c}_i-C_i=\mu \cdot t_0 - \eta \cdot TTC_{fix} \leq 0, 0 \leq \eta \leq 1
\label{eq:constraint definition}
\end{equation}

% add description on geometry
\noindent Given by this constraint value definition, a potential collision revealed in constraint formation will cause the $\bar{c}_i-C_i \geq 0$.

\subsection{Multi-task Reinforcement Learning}

\begin{figure}[htp]
    \centering
    \includegraphics[width=10cm]{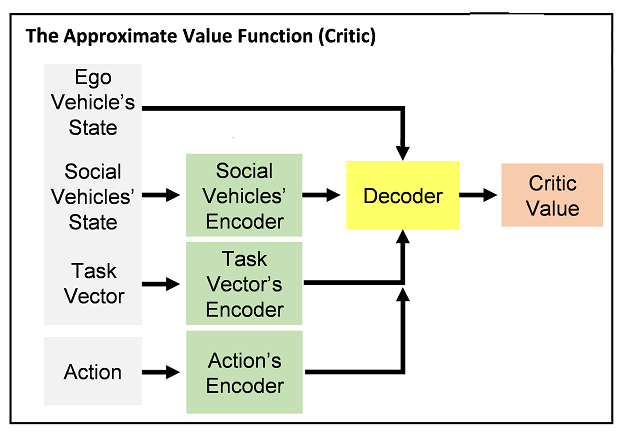}
    \caption{The Structure of the Critic Network}
    \label{fig3}
\end{figure}

As shown in Algorithm \ref{alg1}, our proposed method is composed of safety model and RL model and trained separately. \cite{Kai2020Multi} proposes a multi-task deep Q learning framework which can handle three different unsignalized intersection navigating tasks at the same time. In order to combine with the continuous action safety layer, we select TD3 (Twin Delayed Deep Deterministic policy gradient) \cite{fujimoto2018addressing} algorithm, which is an actor-critic method, to train our multi-task agent. In particular, this technique is mainly used in the approximate value function, namely, critic, whose design is shown in Fig.\ref{fig3}. Note that the encoders and decoder in the critic model are formed by Fully-Connected(FC) layers.

The characteristics of this multi-task RL framework are mainly in three aspects: firstly, all the tasks that agent needs to learn are decomposed as a set of sub-tasks by domain knowledge, i.e. $G=\{g_{s1},…,g_{sn},g_{c1},…g_{cn}\}$, in which $g_{si}$ indicates a specific sub-task and $g_{ci}$ indicates a common sub-task. Secondly, a task code vector which is defined by different combinations of sub-tasks, is included in state representation to indicate the current task of the agent. Thirdly, a vectorized reward value is designed to separate the feedback transmission process of each sub-task to the model.

In the unsignalized intersection navigating task, the sub-tasks set is defined as $G_s= \{ g_{sl},g_{sm},g_{sr},g_c \} $, in which $g_{sl}$, $g_{sm}$ and $g_{sr}$ represent specific sub-task of turning left, going straight and turning right, $g_c$ is a additional dimension to improve the margin of state representation. Along with different combinations of $g_{s \cdot}$ and $g_c$, the task vector $g$ can be defined as a 4-dimension vector, in which the first 3 dimensions imply agent’s current specific sub-task using one-hot code, and the last dimension implies agent’s common sub-task. For example, the task vector is $g=[1,0,0,1]$ for the turning left task,  $g=[0,1,0,1]$ for the going straight task, and $g=[0,0,1,1]$ for turning right task.

Reward is designed as a 4-dimension vector $r=[r_{sl},r_{sm},r_{sr},r_{c}]$ corresponding to the sub-tasks set $G_s$, in which the values of $r_{s \cdot}$ and $r_c$ depend on the performance of sub-task $g_{s \cdot}$ and $g_c$ separately. In this paper, the output of the critic network $Critic(s,a,g)$ is also a 4-dimension vector, which correlates with the design of sub-task set $G_s$ and current task vector $g$. Therefore the state-action value $Q(s,a,g)$ is calculated by the Hadamard product of critic’s output $Critic(s,a,g)$ and task vector $g$
\begin{equation}
    Q(s,a,g)=g^T Critic(s,a,g)
\end{equation} 
The task vector $g$ filters the task-irrelevant values out of $Critic(s,a,g)$, and keeps task-relevant values in $Q(s,a,g)$, which helps to improve the convergence of critic model and prevents the model’s preference for different tasks.

\begin{algorithm}
  \caption{Safe exploration TD3 with Multi-task framework}
  \label{alg1}
  \begin{algorithmic}[1]
  \STATE \textbf{Step.1 Train Safety Model:}
  \STATE Initialize safety layer network $f\left(s ; \omega_i \right)$ with random parameters $\omega_i$, $i\in[K]$  \\
  \STATE Collect datasets $D=\left\{(s,s',a,\bar{c_i},\bar{c_i}')\right\}$ through a random policy. 
  \STATE Train the safety layer $f\left(s ; \omega_i \right)$ using (\ref{eq3})
  \\ \hspace*{\fill} \\
  \STATE \textbf{Step.2 Train Actor and Critic:}
  \STATE Initialize multi-task critic networks $Q_{\phi_1}$, $Q_{\phi_2}$, and social attention actor network $\pi_\theta$ with random parameters $\phi_1$, $\phi_2$,  $\theta$
  \STATE Initialize target networks $\phi_1' \gets \ \phi_1$, $\phi_2' \gets \  \phi_2$, $\theta' \gets \  \theta$
  \STATE Initialize replay buffer $\mathcal{B}$
  \FOR{$t=1$ \textbf{to} $T$}
  \STATE Select action with exploration noise through actor $\hat{a} = \pi(s)+\epsilon$,  $\epsilon \sim \mathcal{N}(0,\psi)$ 
  \STATE Get safe action through safety layer model $a=\hat{a}-\lambda_{i^{*}} f\left(s ; \omega_{i^{*}} \right)$, where $\lambda_{i^{*}}$ is determined by (\ref{eq6})
  \STATE Observe reward $r$ and new state $s'$
  \STATE Store transition tuple $(s,a,r,s')$ in $\mathcal{B}$
  \\ \hspace*{\fill} \\
  \STATE  Sample mini-batch of $N$ transitions $(s,a,r,s')$ from $\mathcal{B}$
  \STATE $\tilde{a} \gets \ \pi_{\theta'}(s)+\epsilon$, $\epsilon \sim$ clip$(\mathcal{N}(0,\tilde{\psi}), -l, l)$ 
  \STATE $q \gets \ r \ + \ \gamma$ $\min_{j=1,2}$ $Q_{\phi_j'}(s', \tilde{a})$
  \STATE Update critics $\phi_j \gets \min_{\phi_j}$ $\frac{1}{N} \sum (q-Q_{\phi_j}(s,a))^{2}$
  \STATE Update $\theta$ by the deterministic policy gradient: \\ $\nabla_\theta J(\theta) = $ $\frac{1}{N} \sum \nabla_a Q_{\phi_1}(s,a) \lvert_{a=\pi_\theta (s)} \nabla_\theta \pi_\theta (s)$
  \STATE Update target networks: \\ $\phi_j' \gets \tau \phi_j + (1-\tau) \phi_j'$ \\ $\theta' \gets \tau \theta + (1-\tau) \theta'$
  \ENDFOR
  \end{algorithmic}
\end{algorithm}

\subsection{Attention Mechanism}

\begin{figure}[h]
    \centering
    \includegraphics[width=12cm]{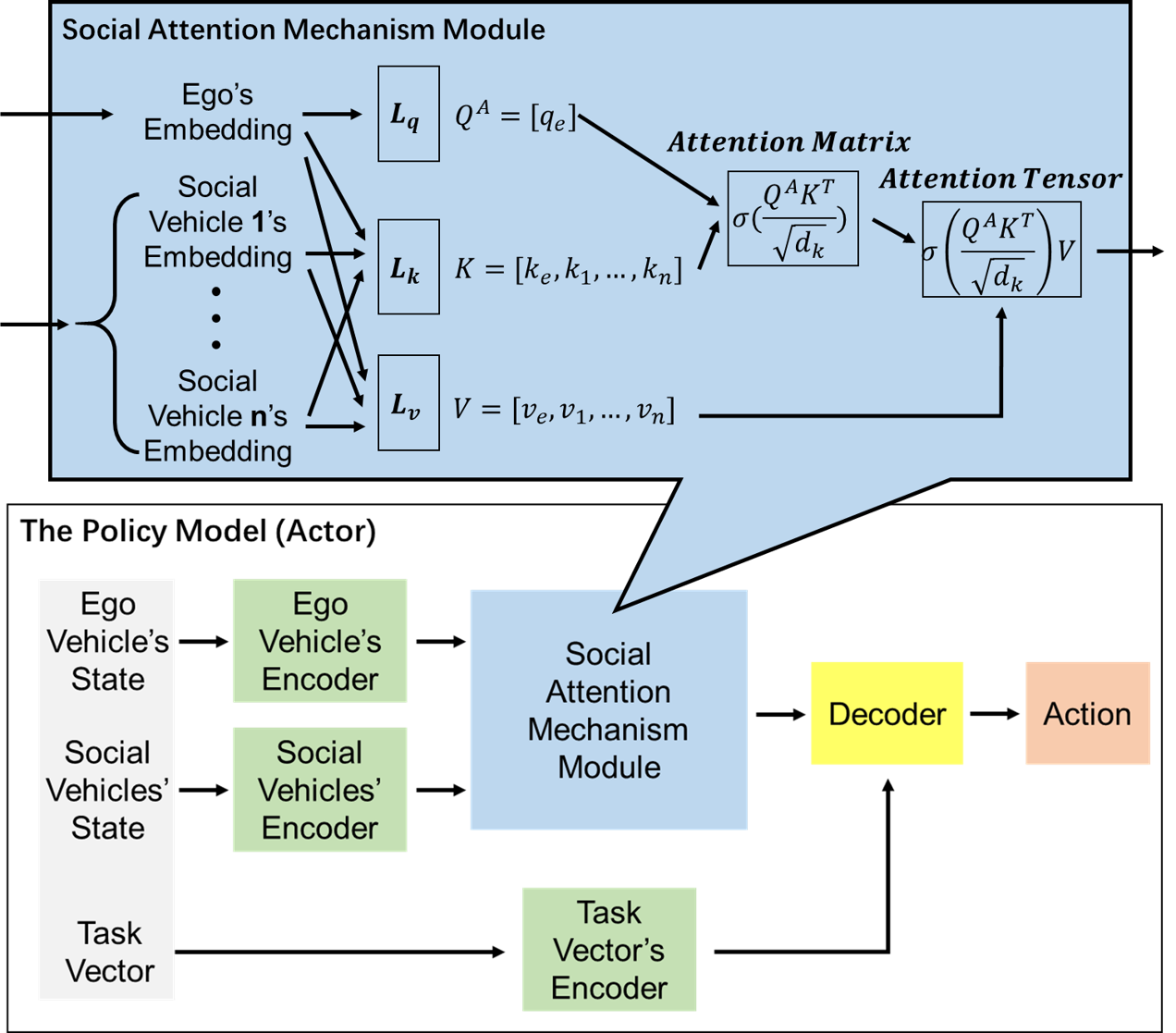}
    \caption{The Structure of the Actor Network}
    \label{Fig: Actor framework}
\end{figure}

The way we employ the attention mechanism to resolve multi-task intersection navigating problem is similar to the social attention mechanism \cite{leurent2019social}. The main purpose of employing this technique is enabling the RL agent to automatically capture dependencies between ego and social vehicles when making a decision, then acquire a better performance as well as better interpretability. In particular, this technique is only used in the policy model, namely, actor. The lower half of Fig.\ref{Fig: Actor framework} shows the structure of the policy model. Obviously, the result of attention mechanism module directly affects the outcome of decision-making. Note that the encoders and decoder in actor model are also formed by Fully-Connected(FC) layers like the critic model.

The process of producing an attention tensor can be described as follows: First, the state representation of all the vehicles need to be decomposed into two parts: the state of ego vehicle and the state of social vehicles, then they are encoded separately and all vehicles’ embeddings are given. The embeddings are fed into the social attention mechanism module, which is shown in the upper half of Fig.\ref{Fig: Actor framework}. 

% \section{Algorithm Name}

There are three nonlinear projections in the module, which are $L_q \in \mathbb{R}^{d_x \times d_k}$, $L_k \in \mathbb{R}^{d_x \times d_k}$, and $L_v \in \mathbb{R}^{d_x \times d_v}$. They are respectively responsible for the generation of query, key, and value vectors. Note that $d_x$ is the length of each vehicle’s embedding, $d_k$ is the length of each query and key vector, $d_v$ is the length of each value vector, and the weights of $L_k$ and $L_v$ are shared between all vehicles. The query vector $Q^A=[q_e] \in \mathbb{R}^{1 \times d_k}$ is calculated by processing the ego vehicle’s embedding with $L_q$, the key vectors $K=[k_e,k_1…k_n] \in \mathbb{R}^{(1+n) \times d_k}$ and the value vectors $V=[v_e,v_1…v_n] \in \mathbb{R}^{(1+n) \times d_v}$ are calculated by processing both ego vehicle’s embedding and social vehicles’ embeddings with $L_k$ and $L_v$. The similarity between the query vector $Q^A$ and the key vectors $K$ can be accessed through their dot product $q_e k_i^T, i \in [e,1,...,n]$. These similarities are then scaled by the inverse square-root-dimension $1/ \sqrt{d_k}$, and normalized with a softmax function $\sigma$ across vehicles, the result stochastic matrix is called $attention \ matrix$, in which the normalized values indicate ego vehicle's attention scores through different traffic participants, including itself. Finally, the product between the $attention \ matrix$ and the value vectors $V$ is the $attention \ tensor$ for forward propagation:
\begin{equation}
    attention \ tensor = \sigma (\frac{Q^AK^T}{\sqrt{d_k}})V
\end{equation} 

\section{Experiments}

In this paper, we deploy our method in two simulation environments, which are developed based on SUMO\cite{SUMO2018} and CARLA\cite{Dosovitskiy17} simulators. The SUMO simulator provides a highly portable interface for intelligent controller deployment, as well as the convenient environmental traffic flow generation. Meanwhile, the CARLA simulator takes a more delicate consideration on the dynamics of vehicles and provides a high-fidelity simulation. In CARLA, we deploy the proposed method into a more realistic autonomous driving pipeline for a further evaluation.

\subsection{SUMO Experiments}

\subsubsection{Experiment Setup}

\begin{figure}[htb]
    \centering
    
    \subfigure[left turn task]{
    \includegraphics[width=6cm]{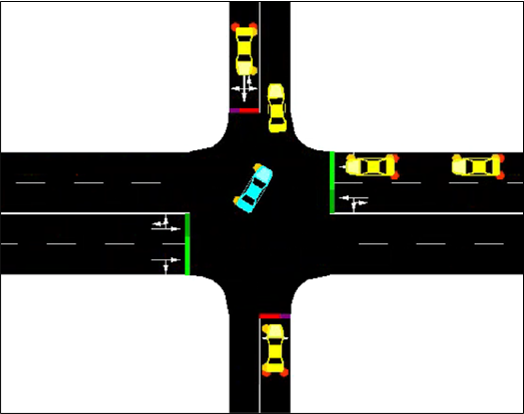}
    %\caption{}
    }
% \quad
    \subfigure[right turn task]{
    \includegraphics[width=6cm]{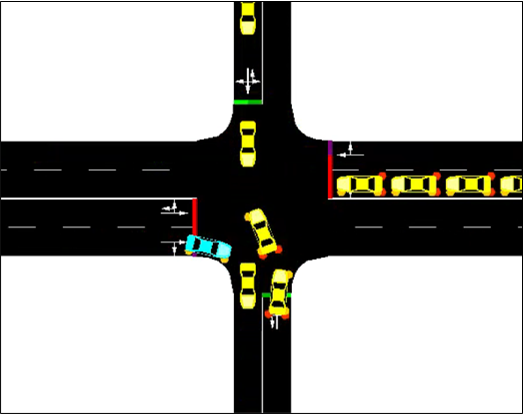}
    %\caption{}
    }
% \subfigure[pic3.]{
% \includegraphics[width=2.5cm]{actor.PNG}
% }
% \quad
% \subfigure[pic4.]{
% \includegraphics[width=2.5cm]{actor.PNG}
% }
\caption{SUMO experiments}
\label{sumo_experiments}
\end{figure}

We employ the SUMO simulator for intersection navigation tasks for clear comparison with related work \cite{Kai2020Multi}, which provides the SOTA performance of the autonomous driving agent for the intersection scenario. An intersection navigation scenario is shown in Fig.\ref{sumo_experiments}, the ego vehicle(cyan) is initially spawn on the west side of the intersection heading east, with north at the top. There is a 4-lane dual carriageway on the east-west direction while a 2-lane dual carriageway on north-south direction.

In SUMO, social vehicles are generated by a continuous traffic flow. We use the same parameters in \cite{Kai2020Multi} in order to compare with the SOTA method. Social vehicles in the traffic are controlled by IDM from SUMO build-in algorithm. Default kinetics parameters are set to both ego and social vehicles so the maximum acceleration will be limited to a realistic value.

% insert training loss figure of safety layer

\subsubsection{Reinforcement Learning Setup}

\textbf{State representations:} We use a 33-dimension vector to represent state information, which can be written as $s = [s_e, s_1,…s_5, g]$. As shown in TABLE \ref{STATE REPRESENTATIONS}, the state vector contains 3 major parts: $s_e$ is a 4-dimension vector referring to the state of ego vehicle, which contains ego vehicle’s speed and a 3-dimension one-hot code, indicating ego vehicle’s current lane. $s_i, i=1,2,...,5$ refer to the state of 5 social vehicles, each of them is a 5-dimension vector which can be written as $s_i=[v_i, x_i, y_i, cos(\alpha_i), sin(\alpha_i)]$, in which $v_i$ indicates social vehicle's speed, $x_i,  y_i$ indicate social vehicle's Cartesian coordinates and $\alpha_i$ indicates social vehicle's heading angle, note that $x_i$, $y_i$ and $\alpha_i$ are all measured under ego vehicle's coordinate system as shown in Fig.\ref{Fig: safety constraint definition}, $g$ is a 4-dimension task vector, where the first 3 dimensions indicate a specific sub-task like going straight, turning left or turning right using one-hot code, and the last dimension indicates a common sub-task like improving traffic efficiency. Note that if the number of social vehicles is larger than 5, only the nearest 5 vehicles are considered, if social vehicles' number is less than 5, then the state vector will be filled to 33-dimension using zero padding.

\begin{table}[h]
\caption{STATE REPRESENTATIONS}
\label{STATE REPRESENTATIONS}
\begin{center}
\begin{tabular}{c|c|c}
\toprule[2pt]
% \quad 
State Component & Description & Feature Length \\
\hline
% \multirow{1}{*}{$s_e$}
$s_e$ & ego vehicle's state & 1*4  \\
% \hline
% \multirow{1}{*}{$s_i$} 
$s_i (i=1,2,...,5)$ & social vehicles' state & 5*5 \\
% \hline
% \multirow{1}{*}{$g$}
$g$ & task vector & 1*4 \\
\hline
% \multirow{1}{*}{$s$}
$s$ & complete state representation & 33 \\
\bottomrule[2pt]
\end{tabular}
\end{center}
\end{table}

In order to select task-relevant social vehicles for the state representation, a filter is designed to reserve vehicles with following rules:
\begin{itemize}
\item vehicle whose distance to ego vehicle is less than a certain threshold (the threshold is 75m in our experiment) will be reserved.
\item vehicles that are in front of ego vehicle. Specifically, as shown in Fig.\ref{Fig: safety constraint definition}, the position vector of a social vehicle in ego coordinate system $\rm {\textbf{F}}_{ego}$ is $R_i=(x_i, y_i)$. The vehicles whose $x_i \geq-5$ will be reserved.
\end{itemize}

\textbf{Action representation:} As we focus on improving autonomous vehicle’s high-level decision-making performance on unsignalized intersection navigating tasks, only longitudinal control of ego vehicle is given by algorithm, the lateral control is assumed to be ideal. 
For policy network, whose output is designed as a 2-dimension vector ${ \left ( a^+, a^-\right )}$. The action of RL module is calculated by $ a = a^+ - a^- $, then normalized to $[0,1]$. Since SUMO simulator uses target speed of vehicle as control command, $a$ is linearly mapped to $(0,9)m/s$ to be transmitted to the simulator. Our experimental result shows that the separate design of the policy network's output can speed up the training process and stabilize RL agent's performance on interacting with other social vehicles.

\textbf{Reward design:} Inspired by \cite{Kai2020Multi,tram2018learning}, the reward value is designed in a vector form. The reward function can be written as $r = [r_{sl}, r_{sm}, r_{sr}, r_c]$, with

\begin{equation}
    r_{s \cdot}=\left\{
    \begin{array}{rcl}
    +30 & , & {current \ task \  success}\\
    -650 & , & {current \  task \ collision}\\
    0 & , & {not \ current \ task}\\
    \end{array} \right. 
    \label{eq: reward definition 1}
\end{equation}

as specific sub-task reward function, which encourages the ego vehicle to reach the target point and punishes the ego vehicle for colliding with social vehicles, and

\begin{equation}
    r_c=\left\{
    \begin{array}{rcl}
    -0.15 & , & {t \leq 0.5 \cdot T_{max} }\\
    -0.3 & , & {t \geq 0.5 \cdot T_{max}}\\
    -50 & , & {time \ exceed}\\
    \end{array} \right.
    \label{eq: reward definition 2}
\end{equation}

where $T_{max}$ refers to maximum time limit of one episode, by which common sub-task reward encourages ego vehicle to improve traffic efficiency.

\subsubsection{Network Architecture}

Generally, the safety model takes state vector of RL model as input and generates a safe action as output. Therefore the input and output dimensions of safety model are the same with RL model, which are 33 and 2 respectively. In this paper, safety model is constructed using neural networks. The actual input of network is the clipped state vector by the following rules. The input layer dimension of safety model for the single task is different from the one used in the multi-task framework. In the single route task, in order to estimate the approximate TTC between ego vehicle and the nearest social vehicle, the original state is clipped for a better convergency. The clipped state consists of the information of ego vehicle and kinetics of nearest social vehicles. Therefore, the first 9 dimensions of the original state vector are used as input of safety layer for the single task. For the multi-task experiment, the 3-dimension sub-task code is supposed to be considered. Therefore we concatenate the 3-dimension one-hot task code with the first 9 dimensions of state vector, constitute a 12-dimension tensor. Besides, the safety model has 3 hidden FC layers with 256 nodes each, connected with the ReLU activation function. Note that the $t_0$ value in (\ref{eq8}) refers to timestep length of the simulation, which is 0.1s in our experiments.

For the RL model, all the encoders and decoders are formed by FC layers with different number of hidden layers and nodes. Vehicle encoder is formed by $64\times64$ FC layers, task encoder is a single FC layer and decoder is formed by $256\times256$ FC layers.

\subsubsection{Results and Analysis}

\paragraph{Training of the Safety Layer} 

\begin{figure}[htb]
\centering
\subfigure[single-task training loss]{
    \includegraphics[width=6cm]{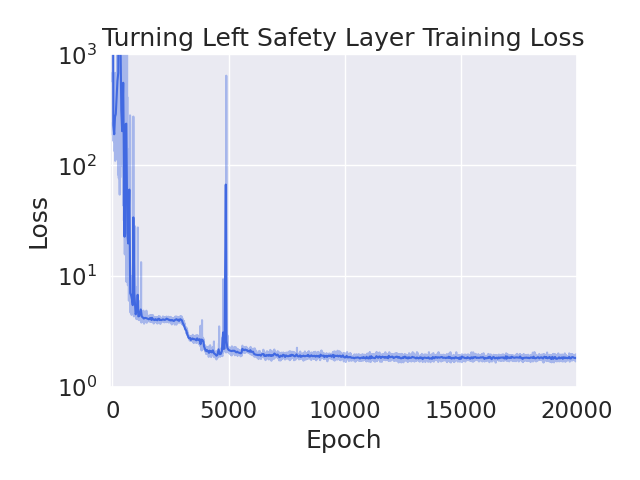}
}
% \quad
\subfigure[multi-task training loss]{
    \includegraphics[width=6cm]{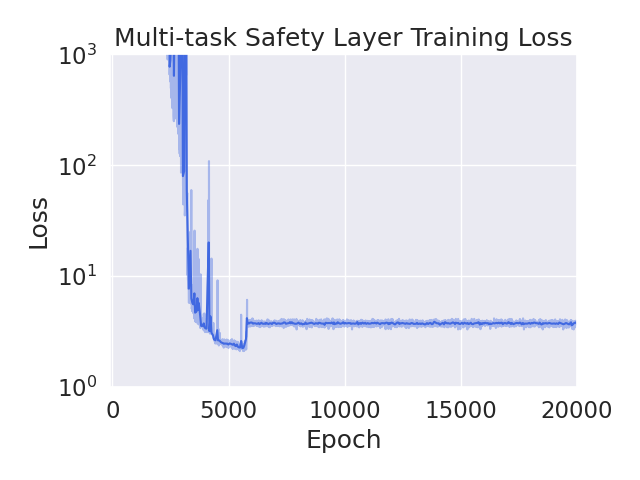}
}

\caption{Training loss of safety layer}
\label{Fig: SUMO safety layer training curve}
\end{figure}

As analyzed in Section \uppercase\expandafter{\romannumeral3}, the definition of safety constraint value is defined using (\ref{eq:constraint definition}). In the SUMO experiments, we set $\mu=1, \eta=0.85$. The safety model is trained using supervised learning. Then the RL agent is trained with the deployment of the safety model to guarantee the safe exploration. Data tuples like ${\left ( s, s', a, \bar c, \bar {c}' \right) }$ are collected through random policy, which means we collect the state vector and constraint value of two continuous timesteps. Normalized posterior error defined by (\ref{eq3}) is used as the training loss. We collect 4 million tuples of data for the training procedure. The training curves of different experiments are shown in Fig.\ref{Fig: SUMO safety layer training curve}.

% insert training loss figure of safety layer

The training results show that the loss values of safety layer for single-task and multi-task eventually converge to a minimum value about 2 and 3, respectively. Since the data is collected by a random policy, the value of the converged loss refers to the minimal estimation error of constraint value.

\paragraph{Performance Metrics}

\begin{figure}[htb]
\centering
\subfigure[single-task rewards]{
    \includegraphics[width=6cm]{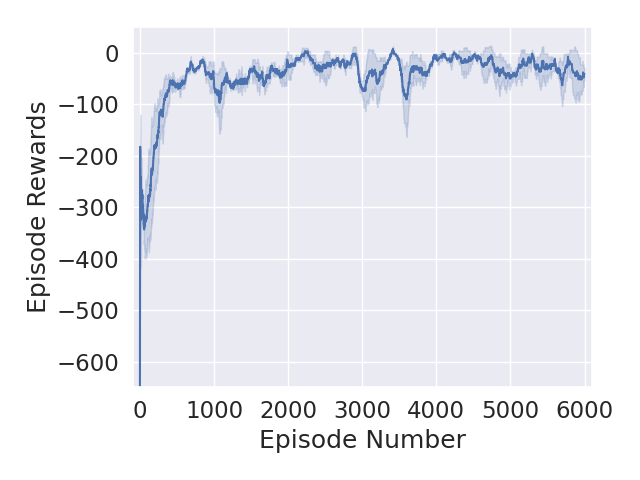}
}
% \quad
\subfigure[single-task success rate]{
    \includegraphics[width=6cm]{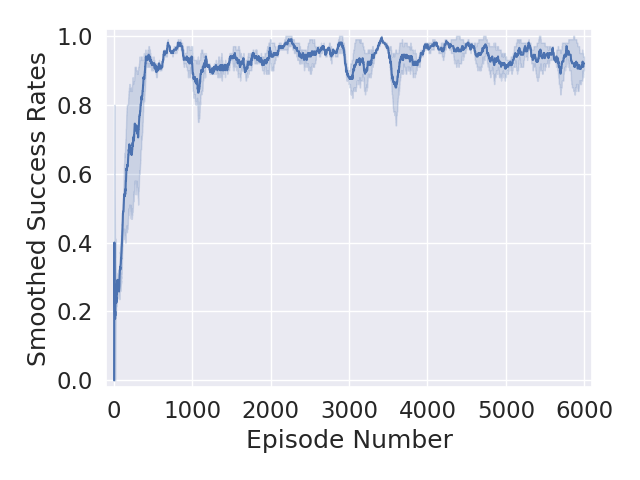}
}
\quad
\subfigure[multi-task rewards]{
    \includegraphics[width=6cm]{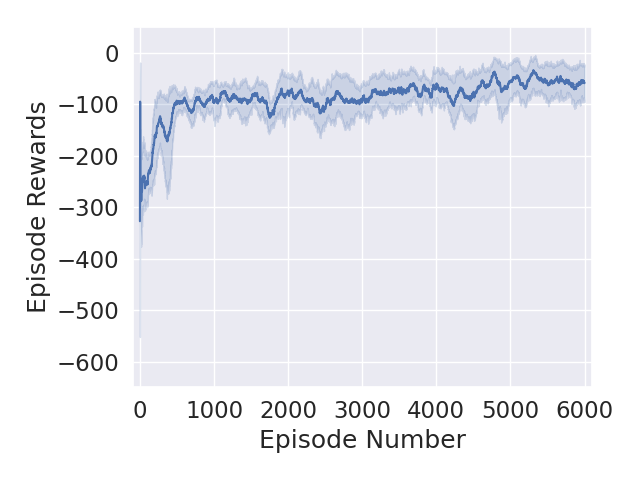}
}
% \quad
\subfigure[multi-task success rate]{
    \includegraphics[width=6cm]{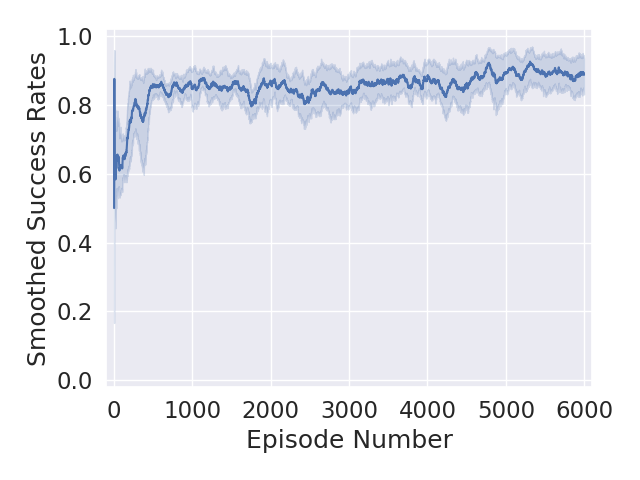}
}
\caption{Single/Multi-task TD3 learning curves}
\label{learning_curves}
\end{figure}

In our experiment, we use TD3 algorithm to train our agent. Two major conditions are compared, the first one is our proposed multi-task TD3 method, which integrates safety layer, social attention mechanism and multi-task framework together. It is trained on three interaction navigation tasks at the same time. The second one is single-task TD3 method which integrates only safety layer and social attention mechanism, which is trained on three tasks separately. Learning curves are shown in Fig.\ref{learning_curves}, in which we compare the success rate and cumulative reward for both conditions. Note that only the most challenging turning left task is compared, since social vehicles will yield to ego vehicle with high probability in the going straight and turning right tasks, which makes it too simple to show significant differences.

% insert training loss figure of safety layer

As Fig.\ref{learning_curves} shows, in turning left task, our proposed multi-task TD3 method reaches a competitive performance against the single-task TD3 method. The convergence of the learning curves also shows that our proposed multi-task TD3 method doesn't have a preference for different tasks and is capable of dealing with a composite unsignalized intersection navigating task.

\begin{figure}[htbp]
\centering
\subfigure[Attention on Ego Vehicle]{
    \includegraphics[width=6cm]{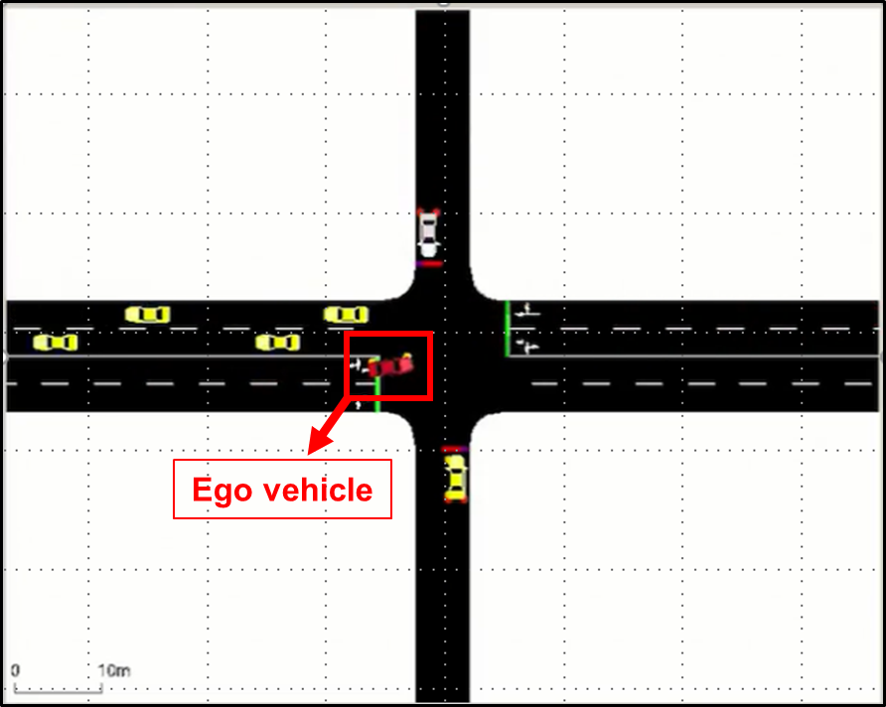}
    \label{Fig: attention_1}
}
% \quad
\subfigure[Attention on Environmental Vehicle]{
    \includegraphics[width=6cm]{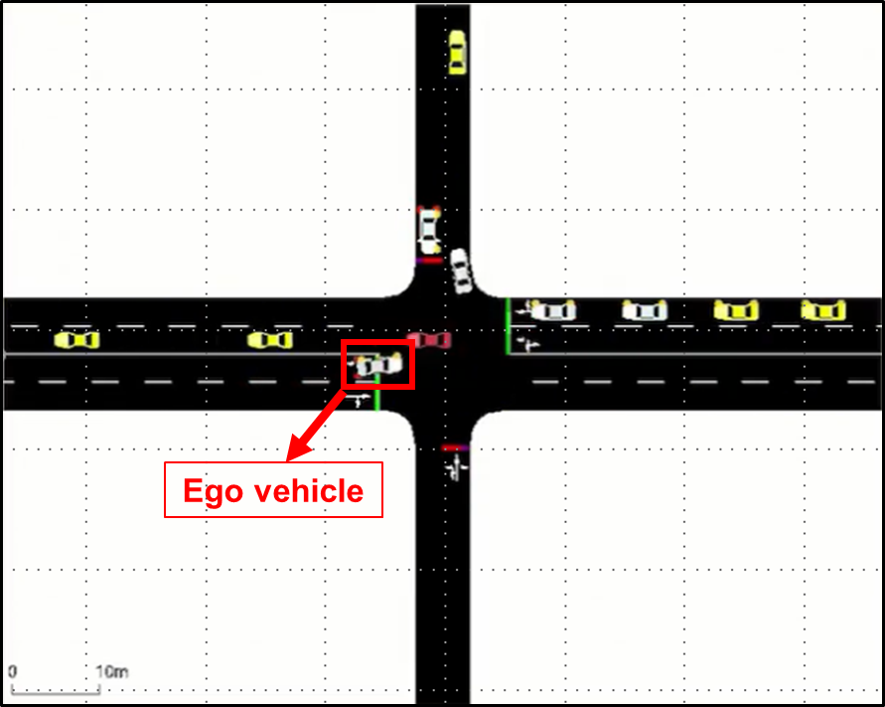}
    \label{Fig: attention_2}
}

\caption{Attention visualization results}
\label{Fig: Attention visualization}
\end{figure}

In the proposed framework, the social attention module is capable of improving the interpretability of the decision-making. The attention weights can be visualized by plotting a color map on the vehicles, as shown in Fig.\ref{Fig: Attention visualization}. In the Fig.\ref{Fig: attention_1}, since the ego vehicle is not interacting with any environmental vehicle, the policy will assign the attention weights on the ego vehicle. Similarly, as shown in Fig.\ref{Fig: attention_2}, the policy network puts attention on the encountering environmental vehicle.

% interpretability of attention

\paragraph{Comparison Results}

To prove the superiority of our proposed method, a comparative study is then employed in the testing setting. Success rate and total episode time is the 2 major indexes used to evaluate the performance of the agent. An agent should finish the route as fast as possible and try to prevent collisions as well. The comparison results which contains 1000 episodes of testing are shown in TABLE \ref{COMPARISON RESULTS}. Our proposed single-task TD3 method outperforms all other methods in turning left task, which has the highest success rate and a relatively less episode time. On the other hand, our proposed multi-task TD3 method shows a competitive result against the single task version, outperforms in traffic efficiency while being with only a slight drop in success rate. Moreover, our proposed multi-task TD3 method fully exceeds the multi-task DQN method from \cite{Kai2020Multi} in both success rate and traffic efficiency: in turning left task, our proposed method increases by 0.2\% in success rate and 24\% in traffic efficiency; in turning right task, our proposed method increases by 3.2\% in success rate and 11.4\% in traffic efficiency.

Besides, a random policy is evaluated as a basic baseline method. IDM is compared as a rule-based baseline method. As the result shows, both single and multi-task methods we proposed exceed the performance of IDM in safety. IDM appears to sacrifice safety to perform well in efficiency, which is not acceptable under most circumstances. However in turning left task, IDM shows a tremendous gap in success rate compared with our proposed method, which shows superiority of our methods in dealing with safe interaction problem.

\begin{table}[htp]
% \resizebox{\textwidth}{!}{
\caption{COMPARISON RESULTS}
\label{COMPARISON RESULTS}
\begin{center}
\begin{tabular}{c|c|c c | c c}
\toprule[2pt]
\multirow{2}{*}{Method} & \multirow{2}{*}{Framework} & \multicolumn{2}{c|}{Success rate(\%) $\uparrow$} & \multicolumn{2}{c}{Average time(s) $\downarrow$} \\
\cline{3-6}
& & left & right & left & right \\
\hline
\rule{0pt}{10pt}
Safety+Attention & single-task & \textbf{98.3} & 99.4 & 10.35 & \textbf{5.35} \\[2pt]
TD3(ours) & multi-task & 97.1 & \textbf{99.5} & 9.40 & 6.62\\[3pt]

% \multirow{2}{*}{}

% & \tabincell{c}{Safe+Attention \\ TD3(ours)}
% & single-task & \textbf{98.3} & 99.4 & 10.35 & \textbf{5.35} \\ 
% & multi-task & 97.1 & \textbf{99.5} & \textbf{9.40} &  6.62\\

\hline
\rule{0pt}{10pt}
\multirow{2}{*}{SOTA DQN\cite{Kai2020Multi}} & single-task & 95.5 & 96.9 & 14.29 & 8.09\\[2pt]
 & multi-task & 96.9 & 96.3 & 12.37 & 7.47 \\[3pt]
 
\hline
\multirow{1}{*}{Random Policy} & \diagbox[width=6em] & 52.4 & 91.2 & 15.96 & 12.46 \\
% \multirow{1}{*}{Random Policy} & / & 52.4 & 91.2 & 15.96 & 12.46 \\
\hline
\multirow{1}{*}{IDM} & \diagbox[width=6em] & 73.4 & 97.3 & \textbf{5.51} & 9.68 \\
\bottomrule[2pt]
\end{tabular}
\end{center}

% }

\end{table}

\paragraph{Ablation Study} 

In order to analyze the impact of the safety layer and attention mechanism on both single-task and multi-task performance more clearly, we have carried out a more detailed experimental analysis. Since the turning left task is the most challenging one, we deploy our ablation study merely in turning left task. The results are shown in TABLE \ref{ABLATION RESULTS}, single-task and multi-task experiments are compared separately, each row in both parts of the table is correlated. The first row is the method deployed with safety layer and attention module, the second and third row refers to TD3 with attention and TD3 with safety respectively, the fourth row refers to TD3 without any additional module. The fifth row refers to a trained TD3 with attention module deployed a pre-trained safety layer in the evaluation phase.

First, we analyze the impact of safety layer. As shown in the first, third and fifth rows of TABLE \ref{ABLATION RESULTS}, the safety layer increases the success rate in single-task and multi-task experiments. On the other hand, attention mechanism is proved to be effective. The attention mechanism significantly reduces the average time of the task, with a slight loss in the success rate in multi-task experiments. And the combination of safety layer and attention module improves success rate and traffic efficiency at the same time in all experiments. 

The major difference is that in single-task condition, attention mechanism outperforms the safety layer in both safety and efficiency, while in multi-task experiments, the safety layer and attention module increase safety and efficiency respectively. We speculate that it is because the multi-task navigation is not conducive to the convergence of the attention module. Finally, we deploy safety layer directly on an agent trained with attention module. The success rate increases without much loss of average time. According to experimental results, the safety layer method significantly improves the safety while not losing much efficiency. Meanwhile, it is obvious that the design of an appropriate safety constraint is critical in the problem.

\begin{table}[htb]
\caption{ABLATION RESULTS}
\label{ABLATION RESULTS}
\begin{center}
 \setlength{\tabcolsep}{2mm}{
\begin{tabular}{c|c| c c }
\toprule[2pt]
Framework & Method & Success rate(\%) $\uparrow$ & Average time(s) $\downarrow$ \\
\hline
\multirow{5}{*}{single-task} 
& TD3 & 92.4 & 12.37 \\
& TD3+Attention  & 94.5 & 13.08 \\
& TD3+Safety & 93.9 & 14.56 \\
& TD3+Safety+Attention & \textbf{98.3} & \textbf{10.35} \\
& pre-trained+Safety & 94.9 & 13.13 \\
\hline
\multirow{5}{*}{multi-task} 
& TD3 & 90.9 & 16.48 \\
& TD3+Attention & 91.2 & \textbf{7.52} \\
& TD3+Safety & 93.2 & 17.04 \\
& TD3+Safety+Attention & \textbf{97.1} & 9.40 \\
& pre-trained+Safety & 91.7 & 7.73 \\
\bottomrule[2pt]
\end{tabular}
}
\end{center}
\end{table}

%\subsection{Discussion}

%According to experimental results, the safety layer method significantly improves the safety while not losing much efficiency. Meanwhile, it is obvious that the design of safety constraint is critical in the problem. At the initial state of the research, environment constraint value is designed using minimum collision radius, which is define by minimum distance between ego vehicle and nearest social vehicle. However such design cause ambiguous constraint value in some certain conditions. For example, before ego vehicle entering into the intersection, the passing social vehicles in opposite direction may reach into the collision radius if the radius value is slacked larger than actual bounding box of ego vehicle. This will bring difficulty in safety constraint value estimation.

\subsection{CARLA Experiments}

Though the SUMO experiments validate the proposed method preliminarily, we wish to further study the effectiveness of safety exploration in a high-fidelity simulator. Compared to SUMO simulator, CARLA simulator\cite{Dosovitskiy17} provides abundant adjustable settings for building a high-fidelity vehicles model. Therefore, we employ the CARLA simulator to build the RL environment. In CARLA experiments, some experimental details are various from the one we used in SUMO experiments, which will be further introduced below.

\subsubsection{Experiment Setup}
\label{subsubsec:carla experiment setup}

\begin{figure}[htp]
    \centering
    \includegraphics[width=8cm]{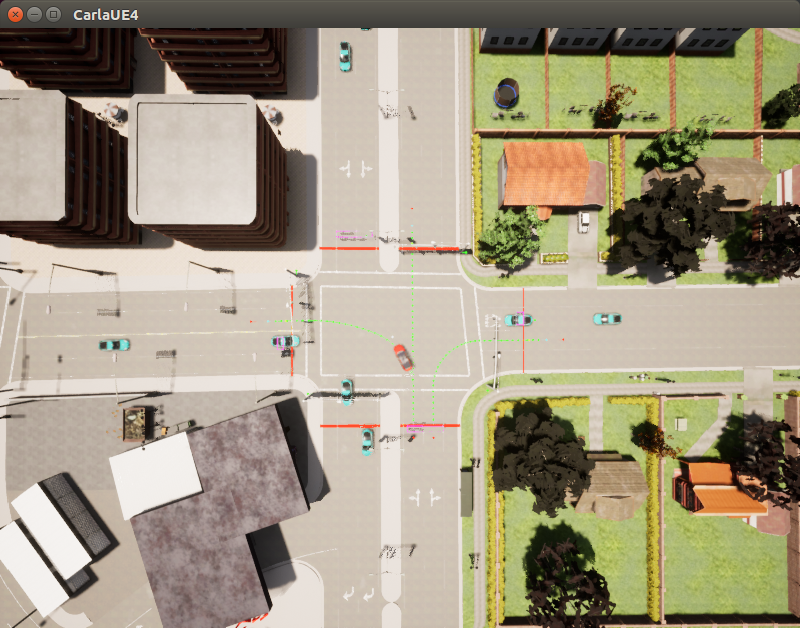}
    \caption{The CARLA experiments}
    \label{Fig: CARLA environment}
\end{figure}

In the CARLA simulator, the Town03 map provides a suitable intersection for the scenario construction, which is shown in Fig.\ref{Fig: CARLA environment}. %Though the road structure is slightly different from the SUMO experiments, the intersection scenario is realistic enough to evaluate the algorithm. 

As shown in the figure, the red bounding box refers to the junction area, and the red vehicle refers to the ego vehicle, which is driven by the RL policy. The target waypoint and the complete route of the ego vehicle are determined according to the task option. All three available routes of the ego vehicle are plotted with green color, as shown in Fig.\ref{Fig: CARLA environment}. For the ego vehicle control, we adopted a common treatment \cite{bouton2019safe} to reduce the dimension of action space. The output of the algorithm is mapped to the target speed of the ego vehicle. A PID controller is designed to obtain the vehicle control command according to the target speed.

Note that the traffic flow in CARLA is slightly different from SUMO experiments. In the CARLA simulator, environmental vehicles are driven by the built-in Autopilot, which provides a set of tunable parameters for the behavior of vehicles. Meanwhile, the routes of all environmental vehicles are determined randomly since the target point of each vehicle controlled by Autopilot is determined recursively and randomly. CARLA provides random seed options to set the behavioral pattern of environmental vehicles implicitly. Meanwhile, the traffic flow behavior in CARLA is determined by a set of critical parameters, including upper-speed limit and collision detection probability(CDP). The CDP indicates the possibility of collision detection between the vehicle and a specific vehicle. We set all the environmental vehicles to fully detect each other, and set the CDP to the ego vehicle as an adjustable parameter. %During the training phase, we setup a parameter decay setting for the CDP value. 

At the beginning of the training, the CDP is set to 0.5, which means each environmental vehicle has a 50\% probability to avoid collision with the ego vehicle. The CDP value will be linearly increased. When the agent is trained after 8000 episodes, the CDP is set to 1.0, which means the environmental vehicles will not detect the collision with the ego vehicle and bring the most challenging situation for the RL training. During the testing phase, the CDP is set to 1.0 permanently to make a solid evaluation.

\subsubsection{Reinforcement Learning Setting}

In CARLA experiments, the RL setting is very similar to the one in the SUMO experiments. The action space definition and the reward function are the same as the SUMO experiments, as shown in (\ref{eq: reward definition 1}) and (\ref{eq: reward definition 2}). For the state representation, we divide the velocity vector into the horizontal plane using two orthogonal vectors, which makes the state vector of each environmental vehicle be a vector of 6 dimensions $s_i=[x_i, y_i, v_{x, i}, v_{y, i}, cos(\alpha_i), sin(\alpha_i)]$. $v_{x, i}$ and $v_{y, i}$ refer to the component of the velocity vector in x and y directions.
In addition, the parameters in (\ref{eq:constraint definition}) are tuned artificially, and be set to $\mu=3, \eta=0.9$. Such settings will keep a larger margin for the safe action correction. In order to make a more understandable deployment of safe action exploration, the safe action modification is only considered when the TTC is less than 4 seconds. Such setup is adopted in both the data collection phase as well as the front propagation phase of the safety layer, including in the RL training and testing.

\subsubsection{Results and Analysis}

\paragraph{Training of the Safety Layer} 

% insert training loss figure of safety layer
\begin{figure}[htb]
    \centering
    \subfigure[single-task training loss]{
        \includegraphics[width=6cm]{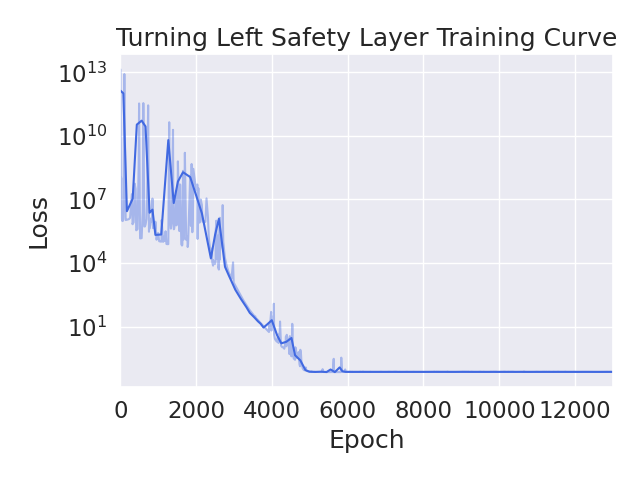}
        %\caption{}
    }
% \quad
    \subfigure[multi-task training loss]{
        \includegraphics[width=6cm]{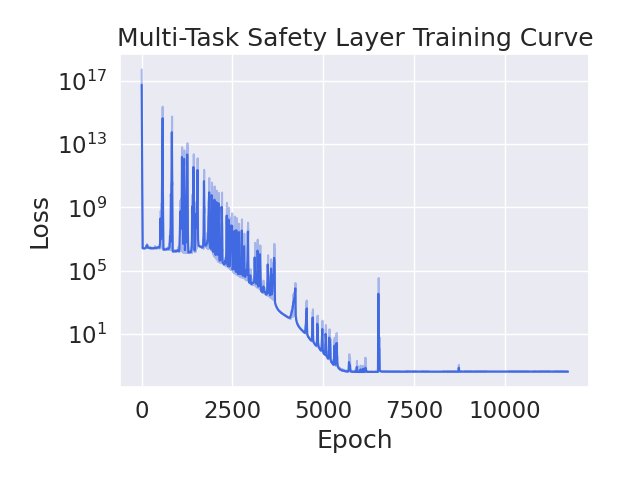}
        %\caption{fig1}
    }
% \quad
% \subfigure[pic3.]{
% \includegraphics[width=2.5cm]{actor.PNG}
% }
% \quad
% \subfigure[pic4.]{
% \includegraphics[width=2.5cm]{actor.PNG}
% }
\caption{Training loss of safety layer}
\label{Fig: CARLA safety layer training curve}
\end{figure}

% illustrate the feature of training curves
The training curves of the safety layer for the turning left task and multi-task are shown in Fig.\ref{Fig: CARLA safety layer training curve}. Since the safety layer is a linear approximation of the safety constraint dynamics, the training of the safety layer network can be seen as a system recognition through supervised learning.  Although there are some fluctuations with the training data collected by a random policy, the training losses can converge to a small error. 

%In the CARLA experiments, some additional assumptions are made to improve the performance of the safety layer module. In the CARLA experiments, the safety constraint is defined by \ref{eq:constraint definition}. In CARLA simulator the inertia of the vehicles is modelled more accurately. 

\paragraph{Performance and Ablation Study}

% insert training loss figure of safety layer
\begin{figure}[htb]
\centering
\subfigure[single-task rewards]{
    \includegraphics[width=6cm]{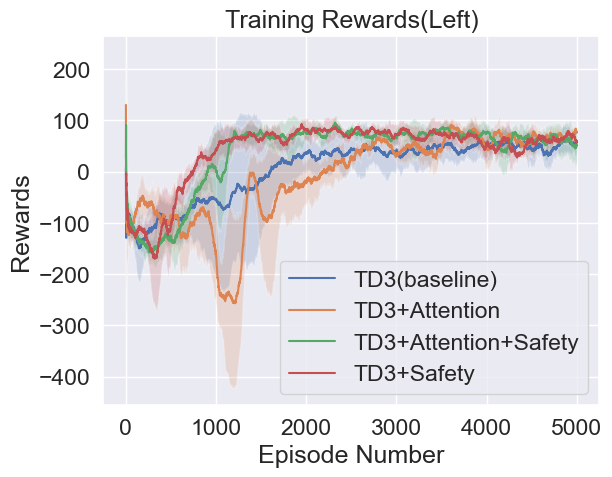}
}
% \quad
\subfigure[single-task success rate]{
    \includegraphics[width=6cm]{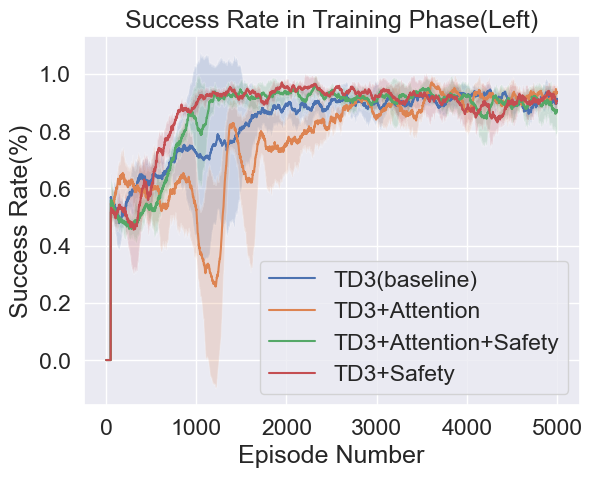}
}

\quad

\subfigure[multi-task rewards]{
    \includegraphics[width=6cm]{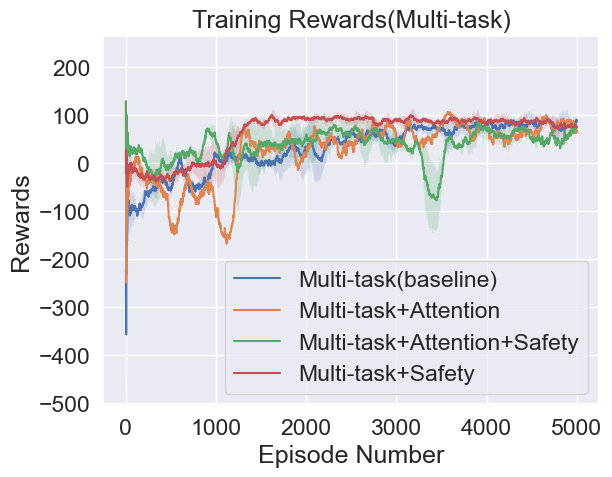}
}
% \quad
\subfigure[multi-task success rate]{
    \includegraphics[width=6cm]{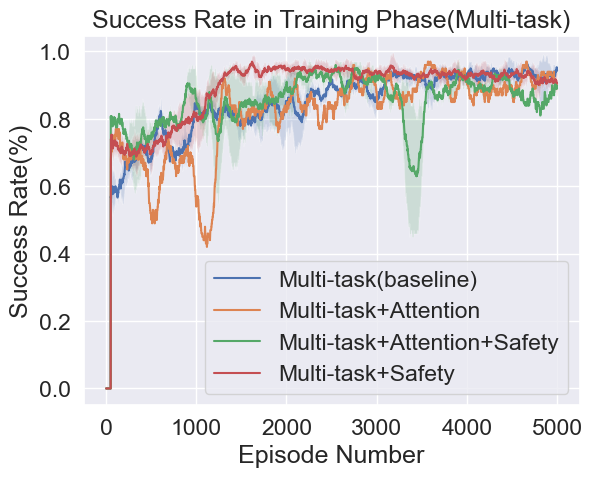}
}
\caption{Single/Multi-task RL learning curves}
\label{Fig: CARLA RL learning curves}
\end{figure}

In CARLA experiments, the performance of the RL agent is evaluated using the same metrics, which are the success rate and average time. %During the testing phase, the ego vehicle is controlled by the trained RL model deployed with the same safety layer model.
% % analysis on TTC and action correction
% In order to make a more detailed analysis on the safe model. We make a statistic on the TTC value and the action modification during the test phase. The action modification and TTC value is counted along a complete episode of the 
The training curves of the CARLA experiments are shown in Fig.\ref{Fig: CARLA RL learning curves}. The training curves are obtained by averaging multiple training sessions. In the CARLA experiments, five training sessions are deployed, and a different random seed value is used in each training session.

From the training curves, we can see that the original TD3 agent is capable of converging to a stable performance. Meanwhile, the TD3 agent which deployed a safety layer slightly outperforms the TD3 baseline. Besides, the introduction of the social attention module brings a significant oscillation on the training process. The social attention mechanism is supposed to emphasize the interpretability of the RL methods. In our experiments, the combination of safety layer and attention module accelerates the converging speed of the training process.

% test results and ablation study
Furthermore, we analyze the ablation performances for the challenging turning left task and multi-task scenarios. In the testing phase, the environmental agents are controlled by the CARLA Autopilot, and the CDP value is set to 1, which brings the most challenging testing situation. According to the results in TABLE \ref{Table: CARLA result}, we can see the safety layer improves the success rate for the random policy, and the TD3 agent deployed with a safety layer reaches better performances than the standard TD3 agents on both single-task and multi-task testing. The agent which combines both the attention module and safety layer reaches the best performance in the single-task testing. Meanwhile, in multi-task testing, such an agent reaches the best success rate in exchange for passing efficiency.

%The ablation study is supposed to make a further validation on the effectiveness of the safety layer.
%Basically, the safety layer improves the success rate for the random policy. And then, the TD3 agent deployed with a safety layer reaches better performances than the standard TD3 agents on both single-task and multi-task tests. The agent which combines both the attention module and safety layer reaches the best performance in the single-task testing. Meanwhile, in the multi-task testing, such an agent reaches the best success rate in exchange for passing efficiency.

\begin{table}[htb]
\caption{CARLA EXPERIMENT RESULTS}
\begin{center}
\label{Table: CARLA result}
 \setlength{\tabcolsep}{2mm}{
\begin{tabular}{c|c| c c }
\toprule[2pt]
Framework & Method & Success rate(\%) $\uparrow$ & Average time(s) $\downarrow$ \\
\hline
\multirow{5}{*}{single-task} 
& Random & 37.5 & 14.5 \\
& Random+Safety & 39.4 & 15.2 \\
& TD3 & 88.6 & 13.0 \\
& TD3+Safety & 89.1 & 11.7 \\
& TD3+Attention  & 81.8 & 11.6 \\
& TD3+Safety+Attention & \textbf{89.6} & \textbf{10.5} \\
% & pre-trained+safety & 94.9 & 13.13 \\
\hline
\multirow{5}{*}{multi-task} 
& Random & 63.1 & 13.4 \\
& Random+Safety & 63.6 & 13.5 \\
& TD3 & 88.1 & 11.2 \\
& TD3+Safety & 88.5 & 8.8 \\
& TD3+Attention & 88.0 & \textbf{8.3} \\
& TD3+Safety+Attention & \textbf{91.2} & 9.64 \\
% & pre-trained+safety & 91.7 & 7.73 \\

\bottomrule[2pt]

\end{tabular}
}
\end{center}
\end{table}

In summary, according to the testing experiments, the effectiveness of the safety layer is proved. The safety layer constrains the action generated by the RL agent and provides an enhancement on safety with a slight sacrifice of efficiency. In our experiments, we also find that the attention mechanism provides a relatively more aggressive policy exploration.

\section{CONCLUSIONS}

In this paper, a multi-task RL framework is proposed to combine attention mechanism and safe exploration with TD3 algorithm. Under such a framework, efficiency and safety are both taken into consideration. A novel design of safety constraint is proposed to represent the collision constraint of the optimization model of the navigation problem. An attention mechanism is also deployed in the framework to improve the interpretability of the algorithm. In order to make an adequate validation for the proposed method, two sets of experiments are deployed in both SUMO and CARLA environments. The SUMO experiment results show that the method achieves very competitive results on the intersection navigation problem, the average time reduces by 24\% while the success rate exceeds by 0.2\% compared with the SOTA method. In the CARLA experiments, it is proved that the safety layer is capable of providing the action modification timely, effectively prevent the collision with environmental vehicles. Our proposed multi-task safe RL framework is capable of dealing with different intersection navigation tasks. The testing environment from CARLA experiments provides a convincible benchmark for the autonomous driving research.

% \appendix

% \section{Sample Appendix Section}
% \label{sec:sample:appendix}
% Lorem ipsum dolor sit amet, consectetur adipiscing elit, sed do eiusmod tempor section \ref{sec:sample1} incididunt ut labore et dolore magna aliqua. Ut enim ad minim veniam, quis nostrud exercitation ullamco laboris nisi ut aliquip ex ea commodo consequat. Duis aute irure dolor in reprehenderit in voluptate velit esse cillum dolore eu fugiat nulla pariatur. Excepteur sint occaecat cupidatat non proident, sunt in culpa qui officia deserunt mollit anim id est laborum.

 \bibliographystyle{elsarticle-num} 
 
%  \bibliography{cas-refs}
 \bibliography{ref}





% \end{thebibliography}
\end{document}